\documentclass[10pt,twocolumn,letterpaper]{article}

\usepackage{ijcb}
\usepackage{times}
\usepackage{epsfig}
\usepackage{graphicx}
\usepackage{amsmath}
\usepackage{amssymb}
\usepackage[norule,symbol,perpage]{footmisc}

\usepackage{multirow}
\usepackage{amssymb}

\ijcbfinalcopy 

\ifijcbfinal\fi

\makeatletter
\def\ps@IEEEtitlepagestyle{
\def\@oddfoot{\mycopyrightnotice}
\def\@evenfoot{}
}
\def\mycopyrightnotice{
{\hfill \footnotesize 978-1-6654-3780-6/21/\$31.00 \copyright 2021 IEEE\hfill}
}
\makeatother

\begin{document}

\title{Iris Presentation Attack Detection by Attention-based and \\ Deep Pixel-wise Binary Supervision Network}

\author{Meiling Fang$^{1,2}$, Naser Damer$^{1,2}$, Fadi Boutros$^{1,2}$,  Florian Kirchbuchner$^{1,2}$, Arjan Kuijper$^{1,2}$\\
$^{1}$Fraunhofer Institute for Computer Graphics Research IGD,
Darmstadt, Germany\\
$^{2}$Mathematical and Applied Visual Computing, TU Darmstadt,
Darmstadt, Germany\\
Email: {meiling.fang@igd.fraunhofer.de}
}

\maketitle
\thispagestyle{empty}

\begin{abstract} 

Iris presentation attack detection (PAD) plays a vital role in iris recognition systems. Most existing CNN-based iris PAD solutions 1) perform only binary label supervision during the training of CNNs, serving global information learning but weakening the capture of local discriminative features, 2) prefer the stacked deeper convolutions or expert-designed networks, raising the risk of overfitting, 3) fuse multiple PAD systems or various types of features, increasing difficulty for deployment on mobile devices. Hence, we propose a novel attention-based deep pixel-wise binary supervision (A-PBS) method. Pixel-wise supervision is first able to capture the fine-grained pixel/patch-level cues. Then, the attention mechanism guides the network to automatically find regions that most contribute to an accurate PAD decision. Extensive experiments are performed on LivDet-Iris 2017 and three other publicly available databases to show the effectiveness and robustness of proposed A-PBS methods. For instance, the A-PBS model achieves an HTER of 6.50\% on the IIITD-WVU database outperforming state-of-the-art methods.
\end{abstract}


\vspace{-5mm}
\section{Introduction} 
\label{sec:int}

In recent years, iris recognition systems are being deployed in many law enforcement or civil applications \cite{DBLP:journals/prl/JainNR16, DBLP:journals/ivc/Boutros20, DBLP:conf/ijcb/Boutros20}. 
However, iris recognition systems are vulnerable to Presentation Attacks (PAs) \cite{livedet17, livdet2020}, performing to obfuscate the identity or impersonate a specific person ranging from printouts, replay, or textured contact lenses. Therefore, Presentation Attack Detection (PAD) field has received increasing attention to secure the recognition systems. 

Recent iris PAD works \cite{crossdomain19, DBLP:conf/fusion/FangDBKK20, DBLP:conf/icb/FangDKK20, DBLP:conf/icb/SharmaR20, DBLP:journals/ivc/FangDBKK21, DBLP:conf/eusipco/FangDKK20} are competing to boost the performance using Convolution Neural Network (CNN) to facilitate discriminative feature learning. Even though the CNN-based algorithms achieved good results under intra-database setups, they do not generalized well across databases and unseen attacks. This situation was verified in the LivDet-Iris competitions. The LivDet-Iris is an international competition series launched in 2013 to assess the current state-of-the-art (SoTA) in the iris PAD field. The two most recent edition took place in 2017 \cite{livedet17} and 2020 \cite{livdet2020}. The results reported in the LivDet-Iris 2017 \cite{livedet17} databases pointed out that there are still advancements to be made in the detection of iris PAs, especially under cross-PA, cross-sensor, or cross-database scenarios. Subsequently, LivDet-Iris 2020 \cite{livdet2020} reported a significant performance degradation on novel PAs, \textcolor{black}{showing that the iris PAD is still a challenging task}. \textcolor{black}{No specific training data was offered and the test data was not publicly available as of now for the LivDet-Iris 2020. }
Therefore, our experiments were conducted on LivDet-Iris 2017 and other three publicly available databases. By reviewing most of the recent iris PAD works (see Sec.~\ref{sec:rw}), we find that all CNN-based iris PAD solutions trained models by binary supervision, i.e., networks were only informed that an iris image is bona fide or attack. Binary supervised training facilitates the use of global information but may cause overfitting. Moreover, the network \textcolor{black}{might
not be able to optimally} locate the regions that contribute the most to make accurate decisions based only on the binary information provided. 

To target these issues, we introduce an Attention-based Pixel-wise Binary Supervision (A-PBS) network (See Fig.\ref{fig:networks}). The main contributions of the work include: 1) we exploit deep Pixel-wise Binary Supervision (PBS) along with binary classification for capturing subtle features in attacking iris samples with the help of spatially positional supervision. 2) we propose a novel effective and robust attention-based PBS (A-PBS) architecture, an extended method of PBS, for fine-grained local information learning in the iris PAD task, 3) we conduct extensive experiments on LivDet-Iris 2017 and other three publicly available databases indicating that our proposed PBS and A-PBS solution outperforms SoTA PAD approaches in most experimental settings. Moreover, the A-PBS method exhibits generalizability across unknown PAs, sensors, and databases.

\vspace{-3mm}
\section{Related Works} 
\label{sec:rw}

\textbf{CNN-based iris PAD:} In recent years, many works \cite{DBLP:conf/icb/FangDKK20, DBLP:conf/fusion/FangDBKK20, DBLP:conf/icb/SharmaR20, crossdomain19, fusionvgg18, DBLP:conf/eusipco/FangDKK20} leveraged deep learning techniques and showed great progress in iris PAD performance. Kuehlkamp \etal \cite{crossdomain19} explored combinations of CNNs with the hand-crafted features. 
However, training $61$ CNNs needs high computational resources and can be considered as an over-tailored solution. Yadav \etal \cite{fusionvgg18} also employed the fusion of hand-crafted features with CNN features and achieved good results. 
Unlike fusing the hand-crafted and CNN-based features, Fang \etal \cite{DBLP:conf/fusion/FangDBKK20} suggested a multi-layer deep features fusion approach (MLF) based on the characteristics of networks that different convolution layers encode the different levels of information. Apart from the fusion methods, a deep learning-based framework named Micro Stripe Analyses (MSA) \cite{DBLP:conf/icb/FangDKK20, DBLP:journals/ivc/FangDBKK21} was introduce to capture the artifacts around the iris/sclera boundary and showed a good performance on textured lens attacks. Yadav \etal \cite{densepad19} presented DensePAD method to detec PAs by utilizing DenseNet architecture \cite{densenet}. 
Furthermore, Sharma and Ross \cite{DBLP:conf/icb/SharmaR20} also exploited the architectural benefits of DenseNet \cite{densenet} to propose an iris PA detector (D-NetPAD) evaluated on a proprietary database and the LivDet-Iris 2017 databases. 
Although fine-tuned D-NetPAD achieved good results on LivDet-Iris 2017 databases with the help of their private additional data, scratch D-NetPAD still failed in the case of cross-database scenarios. 
These works inspired us to use DenseNet \cite{densenet} as the backbone for further design of network architectures.
\textcolor{black}{Very recently, Chen \etal \cite{DBLP:conf/wacv/ChenR21} proposed an attention-guided iris PAD method for refine the feature maps of DenseNet \cite{densenet}. However, this work used conventional sample binary supervision and did not report cross-database experiments to verify the generalizability of the added attention module. } 

\textbf{Limitations:} To our knowledge, current CNN-based iris PAD solutions train models only through binary supervision (bona fide or attack). From the above recent iris PAD literature, it can be seen that deep learning-based methods boost the performance but still have the risk of overfitting under cross-PA/cross-database scenarios. Hence, some recent methods proposed the fusion of multiple PAD systems or features to improve the generalizability, which makes it challenging for deployment. One of the major reasons causing overfitting is the lack of availability of a sufficient amount of variant iris data for training networks. Another possible reason might be binary supervision. While the binary classification model provides useful global information, its ability to capture subtle differences in attacking iris samples may be weakened, and thus the deep features might be less discriminative. This possible cause motivates us to exploit binary masks to supervise the training of our PAD model, because a binary mask label may help to supervise the information at each spatial location. However, PBS may also lead to another issue, as the model misses the exploration of important regions due to the 'equal' focus on each pixel/patch. To overcome some of these difficulties, we propose the A-PBS architecture to force the network to find regions that should be emphasized or suppressed for a more accurate iris PAD decision. The detailed introduction of PBS and A-PBS can be found in Sec.~\ref{sec:meh}.

\vspace{-3mm}
\section{Methodology} 
\label{sec:meh}

\begin{figure*}[htbp!]
\centering{\includegraphics[width=0.9\linewidth]{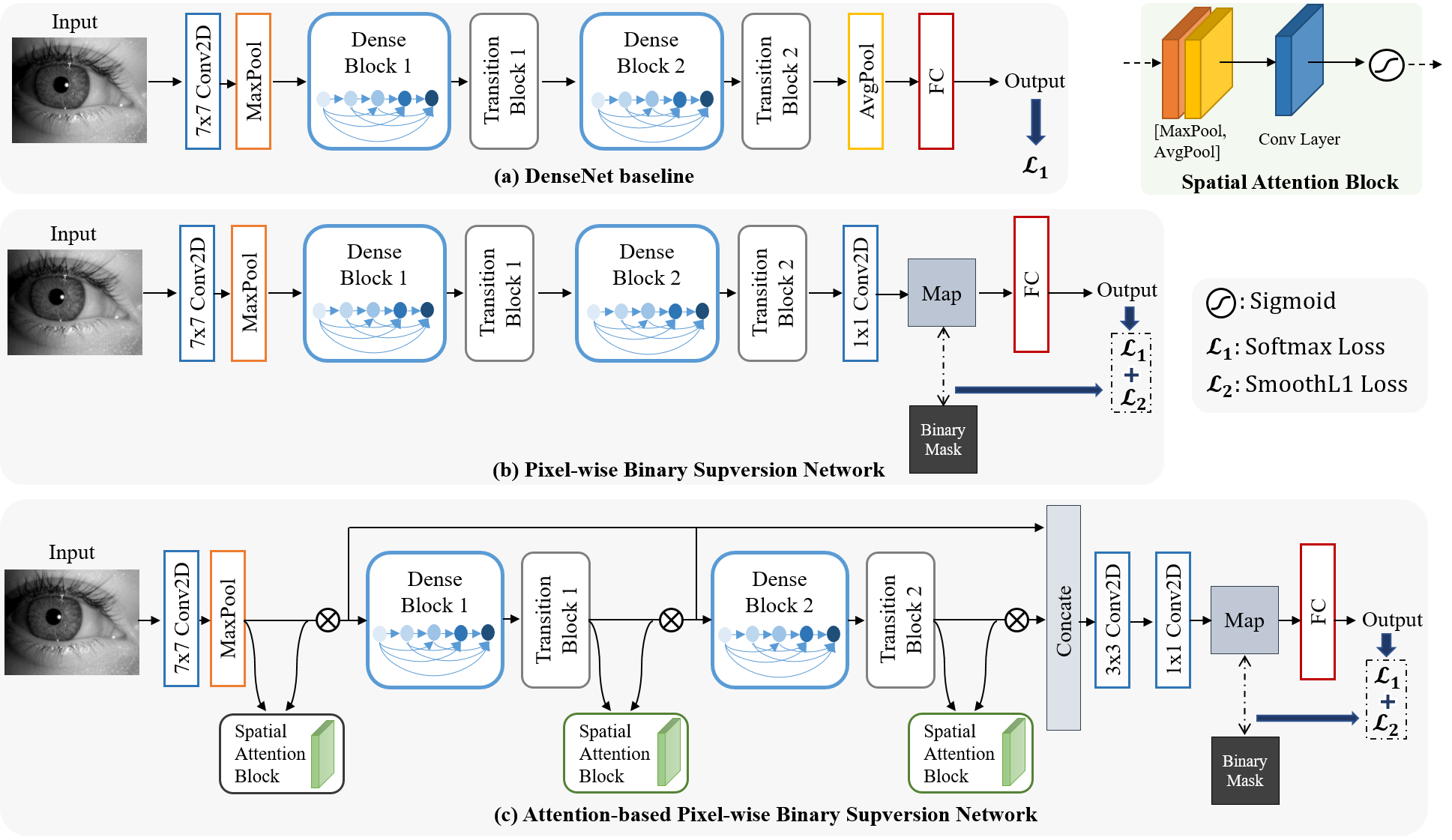}}
\caption{An overview of (a) baseline DenseNet, (b) proposed PBS and (c) proposed A-PBS networks. \label{fig:networks}}
\vspace{-4mm}
\end{figure*}

In this section, we first introduce DenseNet \cite{densenet} as a preliminary architecture. Then, our proposed Pixel-wise Binary Supervision (PBS) and Attention-based PBS (A-PBS) methods are described. As shown in Fig.~\ref{fig:networks}, the first gray block (a) presents the basic DenseNet architecture with binary supervision, the second gray block (b) introduces the binary and PBS, and the third block (c) is the PBS with the fused multi-scale spatial attention mechanism (A-PBS).
\vspace{-2mm}
\subsection{Baseline: DenseNet}
DenseNet \cite{densenet} introduced direct connection between any two layers with the same feature-map size in a feed-forward fashion. 
The reasons motivating our choice of DensetNet are: 1) whilst following a simple connectivity rule, DenseNets naturally integrate the properties of identity mappings and deep supervision. 2) DenseNet has already shown its superiority in ris PAD \cite{densepad19, DBLP:conf/icb/SharmaR20, livdet2020}. As shown in Fig.~\ref{fig:networks}.(a), we reuse two dense and transition blocks of pre-trained DenseNet121. Following the second transition block, an average pooling layer and a fully-connected (FC) classification layer are sequentially appended to generate the final prediction to determine whether the iris image is bona fide or attack. PBS and A-PBS networks will be expanded on this basic architecture later.

\vspace{-2mm}
\subsection{Pixel-wise Binary Supervision Network (PBS)}
From the recent iris PAD literature \cite{DBLP:conf/icb/FangDKK20, DBLP:conf/fusion/FangDBKK20, DBLP:conf/icb/SharmaR20, crossdomain19}, it can be seen that CNN-based methods outperformed hand-crafted feature-based methods. In typical CNN-based iris PAD methods, networks are designed such that feeding pre-processed iris image as input to learn discriminative features between bona fide and artifacts. To that end, a FC layer is generally introduced to output a prediction score supervised by binary label (bona fide or attack). Recent face PAD works have shown that auxiliary supervision \cite{DBLP:conf/cvpr/LiuJ018, deeppixbis} improved their performance. Binary label supervised classification learns semantic features by capture global information but may cause overfitting. Moreover, such embedded 'globally' features might lose the local detailed information in spatial position. These drawbacks give us the insight that additional pixel-wise binary along with binary supervision might improve the iris attack detection results. First, such supervision approach can be seen as a combination of patch-based and vanilla CNN based methods. To be specific, each pixel-wise score in output feature map is considered as the score generated from the patches in an iris image. Second, the binary mask supervision would be provided for the deep embedding features in each spatial position. Hence, an intermediate feature map is predicted before the final binary classification layer (as shown in Fig.~\ref{fig:networks}.(b)). The output from the \textit{Transition Block 2} is 384 channels with the size of $14 \times 14$. A $1 \times 1$ convolution layer is added to produced the map. Finally, a FC layer is used to generates prediction.

\vspace{-2mm}
\subsection{Attention-based PBS Network (A-PBS)}
The architecture of PBS is designed coarsely (simply utilizing the intermediate feature map) based on the DenseNet \cite{densenet}, which might be sub-optimal for iris PAD task. To enhance that, and inspired by Convolutional Block Attention Mechanism (CBAM) \cite{cbam} and MLF \cite{DBLP:conf/fusion/FangDBKK20}, we propose an A-PBS method with multi-scale feature fusion (Fig.~\ref{fig:networks}.(c)). 

Although PBS can boost performance of iris PAD, it shows imperfect invariation under more complicated cross-PA or cross-database scenarios (See results in Tab.~\ref{tab:cross_db}). As a result, it is worth finding the important regions to focus on, although it contradicts learning \textit{more} discriminative features. In contrast, the attention mechanism aims to automatically learn \textit{essential} discriminate features from inputs that are relevant to attack detection. Woo \etal \cite{cbam} presented an attention module consisting of the channel and spatial distinctive sub-modules, which possessed consistent improvements in classification and detection performance with various network architectures. However, only spatial attention module is adopted in our case due to the following reasons: 1) the Squeeze-and-Excitation (SE) based channel attention module focuses only on the inter-channel relationship by using dedicated global feature descriptors, which leads to a loss of information (e.g., class-deterministic pixels) and may result in further performance degradation when the domain is shifted, e.g., different sensors and changing illumination, 2) the spatial attention module utilizes the inter-spatial relationship of features. Specifically, it focuses on \textit{'where'} is an informative part, which is more proper for producing output feature maps for supervision. Moreover, based on the fact that the network embeds different layers of information at different levels of abstraction, the MLF \cite{DBLP:conf/fusion/FangDBKK20} approach confirmed that the fusing deep feature from multiple layers is beneficial to enhance the robustness of the networks in the iris PAD task. Nevertheless, we propose to fuse feature maps generated from different levels directly inside the network instead of fusing features extracted from a trained model in MLF \cite{DBLP:conf/fusion/FangDBKK20}. One reason is that finding the best combination of network layers to fuse is a challenging task and difficult to generalize well, especially when targeting different network architectures.

As illustrated in Fig.~\ref{fig:networks}, three spatial attention modules are added after \textit{MaxPool}, \textit{Transition Block 1}, and \textit{Transition Block 2}, respectively. The feature learned from the \textit{MaxPool} or two \textit{Transition Blocks} can be considered as low-, middle- and high-level features and denoted as \[\mathcal{F}_{level} \in \mathbb{R}^{C \times H \times W}, \quad \textit{level} \in \textit{ \{low, mid, high\} }\]. Then, the generated attention maps $\mathcal{A}_{level} \in \mathbb{R}^{H \times W} $ encoding where to emphasize or suppress are used to refine $\mathcal{F}_{level}$. The refined feature $\mathcal{F}'_{level}$can be formulated as $\mathcal{F}'_{level} = \mathcal{F}_{level} \otimes \mathcal{A}_{level}$ where $\otimes$ is matrix multiplication. Finally, such three different level refined features are concatenated together and then fed into a $1 \times 1$ convolution layer to produce the pixel-wise feature map for supervision. It should be noticed that the size of convolutional kernel in three spatial attention modules is different. As mentioned earlier, the deeper the network layer, the more complex and abstract the extracted features. Therefore, we should use smaller convolutional kernels for deeper features to locate useful region. The kernel sizes of low-, middle- and high-level layers are thus set to 7, 5, and 3, respectively. The experiments have been demonstrated later in Sec.~\ref{sec:es} and showed that A-PBS network possesses better performance and generalizability than the PBS approach.  

\vspace{-2mm}
\subsection{Loss Function}
For the loss function, Binary Cross Entropy (BCE) loss is used for final binary supervision. For the sake of robust PBS needed in iris PAD, Smooth L1 (SmoothL1) loss is utilized to help the network reduce its sensitivity to outliers in the feature map. The equations for SmoothL1 is shown below:
\[
\mathcal{L}_{SmoothL1} = \dfrac{1}{n} \sum z
\]
\[
\indent \indent \text{where} \quad z = 
\begin{cases}
\dfrac{1}{2}\cdot (y - x)^{2}, & \textit{if} \quad |y - x| < 1 \\
|y-x| - \dfrac{1}{2}, & \textit{otherwise}
\end{cases}
\]
where $n$ is the amount number of pixels in the output map ($14 \times 14$ in our case). The equation of BCE is:
\[
\mathcal{L}_{BCE} = -[y \cdot \log p + (1-y)\cdot \log (1-p)]
\]
where $y$ in both loss equations presents the ground truth label. $x$ in SmoothL1 loss presents to the value in feature map, while $p$ in BCE loss is predicted probability.
The overall loss $\mathcal{L}_{overall}$ is formulated as $\mathcal{L}_{overall} = \lambda \cdot \mathcal{L}_{SmoothL1} + (1 - \lambda) \cdot \mathcal{L}_{BCE}$. The $\lambda$ is set to 0.2 in our experiments.

\vspace{-3mm}
\subsection{Implementation Details}
For the databases, whose distribution of bona fides and attacks are imbalanced in the training set, class balancing is done by under-sampling the majority class. Data augmentation is performed during training using random horizontal flips with a probability of 0.5. By considering the limited amount of iris data, the model weight of DenseNet, PBS and A-PBS models are first initialized by the base architecture DenseNet121 trained on the ImageNet dataset and then fine-tuned by iris PAD data. The Adam optimizer is used for training with a initial learning rate of $1e^{-4}$ and a weight decay of $1e^{-6}$. To avoid overfitting, the model is trained with the maximum 20 epochs and the learning rate halved every 6 epochs. The batch size is 64. In the testing stage, we use the binary output as the final prediction score. The proposed method was implemented using the Pytorch.
\vspace{-3mm}
\section{Experimental Evaluation} 
\label{sec:es}

\subsection{Databases}
\label{ssec:db}

\begin{table*}[thbp!]
\begin{center}
\resizebox{0.85\textwidth}{!}{%
\begin{tabular}{|l|l|l|l|l|}
\hline
\multicolumn{2}{|l|}{Database} & \# Training & \# Testing & Type of Iris Images   \\ \hline
\multicolumn{2}{|l|}{NDCLD-2015 \cite{ndcld15}}  & 6,000 & 1,300  & Real, soft and textured lens  \\ \hline
\multirow{2}{*}{NDCLD-2013 \cite{ndcld2013}}  & LG4000 & 3,000 & 1,200  & Real, soft and textured lens   \\ \cline{2-5} 
& AD100  & 600 & 300 & Real, soft and textured lens   \\ \hline
\multirow{2}{*}{IIIT-D CLI \cite{iiitd_cli_2, iiitd_cli}} & Cognet & 1,723 & 1,785 & Real, soft and textured lens   \\ \cline{2-5} 
& Vista  & 1,523 & 1,553 & Real, soft and textured lens    \\ \hline
\multirow{2}{*}{LivDet-Iris 2017 \cite{livedet17}} & Clarkson (cross-PAD) & 4937 & 3158 & Real, textured lens, printouts \\ \cline{2-5}
& Notre Dame (cross-PA) & 1,200 & 3,600 &  Real, textured lenses  \\ \cline{2-5}
& IIITD-WVU (cross-DB) & 6,250 & 4,209 & Real, textured lenses, printouts, lens printouts \\ \hline
\end{tabular}}
\end{center}
\caption{Characteristics of the used databases. All databases have the training and test sets based on their own experimental setting in the related papers. The Warsaw database in Iris-LivDet-2017 competition are no longer publicly available.}
\label{Tab:db_description}
\vspace{-4mm}
\end{table*}

The proposed method is evaluated on multiple databases: three databases comprising of textured contact lens attacks captured by different sensors \cite{ndcld15, ndcld2013, iiitd_cli}, and three databases (Clarkson, Notre Dame and IIITD-WVU) from the LivDet-Iris 2017 competition \cite{livedet17}. The Warsaw database in the LivDet-Iris 2017 is no longer publicly available due to General Data Protection Regulation (GDPR) issues. For the experiments in NDCLD13, NDCLD15, IIIT-CLI databases, 5-fold cross-validation was performed due to no pre-defined training and testing sets. For the experiments in competition databases, we followed the defined data partition and experimental setting \cite{livedet17}. Subjects in each fold or defined partition are dis-joint. The summery of the used databases is listed in Tab~\ref{Tab:db_description}. 

\textbf{NDCLD13:} The NDCLD13 database consists of $5100$ images and is conceptually divided into two sets: 1) LG4000 including $4200$ images captured by IrisAccess LG4000 camera, 2) AD100 comprising of $900$ images captured by risGuard AD100 camera. Both the training and the test set are divided equally into no lens (bona fide), soft lens (bona fide), and textured lens (attack) classes.

\textbf{NDCLD15:} The $7300$ images in the NDCLD15 \cite{ndcld15} were captured by two sensors, IrisGuard AD100 and IrisAccess LG4000 under MIR illumination and controlled environments. The NDCLD15 contains iris images wearing no lenses, soft lenses, textured lenses.

\textbf{IIIT-D CLI:} IIIT-D CLI database contains $6570$ iris images of $101$ subjects with left and right eyes. For each individual, three types of images were captured: 1) no lens, 2) soft lens, and 3) textured lens. Iris images are divided into two sets based on captured sensors: 1) Cogent dual iris sensor and 2) VistaFA2E single iris sensor.

\textbf{LivDet-Iris 2017 Database:} Though the new edition LivDet-Iris competition was held in 2020, we still evaluate the algorithms in databases provided by LivDet-Iris 2017 for several reasons. First, no official training data was announced in the LivDet-Iris 2020, because the organizers encouraged the participants to use all available data (both publicly and proprietary) to improve the effectiveness and robustness. Second, the test data is not publicly available. To make a fair comparison with SoTA algorithms on equivalent data, we use databases in LivDet-Iris 2017 for restricting the factors affecting the evaluation to the algorithm itself rather than the data. Third, the LivDet-Iris 2017 competition databases are still challenging due to the cross-PA and cross-database scenario settings.  
The Clarkson and Notre Dame database are designed for cross-PA scenarios, while the IIIT-WVU database is designed for a cross-database evaluation due to the different sensors and acquisition environments. The Clarkson testing set includes additional unknown visible light image printouts and unknown textured lenses (unknown pattern). Moreover, Notre Dame focused on the unknown textured lenses. However, the Warsaw database is no longer publicly available. 
\vspace{-2mm}
\subsection{Evaluation Metrics}

The following metrics are used to measure the PAD algorithm performance: 1) Attack Presentation Classification Error Rate (APCER), the proportion of attack images incorrectly classified as bona fide samples, 2) Bona fide Presentation Classification Error Rate (BPCER), the proportion of bona fide images incorrectly classified as attack samples, 3) Half Total Error Rate (HTER), the average of APCER and BPCER. The APCER and BPCER follows the standard definition presented in the ISO/IEC 30107-3 \cite{ISO301073} and adopted in most PAD literature including in LivDet-Iris 2017. The threshold for determining the APCER and BPCER is 0.5 as defined in the LivDet-Iris 2017 protocol. In addition, for further comparison on IIITD-CLI \cite{iiitd_cli, iiitd_cli_2} database, we also report the Correct Classification Accuracy (CCR), which is the ratio between the total number of correctly classified images and the number of all classified presentations. Moreover, the performance of our proposed methods is evaluated in terms of True Detection Rate (TDR) at a False Detection Rate (FDR) of 0.2\% (TDR at 0.2\% FDR is normally used to demonstrate the PAD performance in practice). TDR is 1 -APCER, and FDR is the same as BPCER. An Equal Error Rate (EER) locating at the intersection of APCER and BPCER is also reported in Tab.~\ref{tab:cross_db}. The metrics beyond APCER and BPCER are presented to enable a direct comparison with reported results in SoTAs.  


\vspace{-2mm}
\subsection{Results on LivDet-Iris 2017 Database}

\begin{table*}[htbp!]
\begin{center}
\resizebox{0.9\textwidth}{!}{%
\begin{tabular}{|c|c|c|c|c|c|c|c||c|c|c|}
\hline
\multirow{2}{*}{Database} & \multirow{2}{*}{Metric} & \multicolumn{9}{c|}{ PAD Algorithm (\%)} \\ \cline{3-11} 
 &  & Winner \cite{livedet17} & SpoofNet \cite{spoofnet_tuning} & Meta-Fusion \cite{crossdomain19} & D-NetPAD \cite{DBLP:conf/icb/SharmaR20} & MLF \cite{DBLP:conf/fusion/FangDBKK20} & MSA \cite{DBLP:conf/icb/FangDKK20, DBLP:journals/ivc/FangDBKK21} & DenseNet & PBS & A-PBS \\ \hline
\multirow{3}{*}{Clarkson} & APCER & 13.39 & 33.00 & 18.66 & 5.78 & - & - & 10.64 & 8.97 & 6.16 \\ 
 & BPCER & 0.81 & 0.00 & 0.24 & 0.94 & - & - & 0.00 & 0.00 & 0.81 \\ 
 & HTER & 7.10 & 16.50 & 9.45 & \textbf{3.36} & - & - & 5.32 & 4.48 & \textbf{3.48} \\ \hline \hline
\multirow{3}{*}{Notre Dame} & APCER & 7.78 & 18.05 & 4.61 & 10.38 & 2.71 & 12.28 & 16.00 & 8.89 & 7.88 \\
 & BPCER & 0.28 & 0.94 & 1.94 & 3.32 & 1.89 & 0.17 & 0.28 & 1.06 & 0.00 \\ 
 & HTER & 4.03 & 9.50 & \textbf{3.28} & 6.81 & \textbf{2.31} & 6.23 & 8.14 & 4.97 & 3.94 \\ \hline \hline
\multirow{3}{*}{IIITD-WVU} & APCER & 29.40 & 0.34 & 12.32 & 36.41 & 5.39 & 2.31 & 2.88 & 5.76 & 8.86 \\
 & BPCER & 3.99 & 36.89 & 17.52 & 10.12 & 24.79 & 19.94 & 17.95 & 8.26 & 4.13 \\
 & HTER & 16.70 & 18.62 & 14.92 & 23.27 & 15.09 & 11.13 & 10.41 & \textbf{7.01} & \textbf{6.50} \\ \hline
\end{tabular}}
\end{center}
\caption{Iris PAD performance of our proposed methods and existing SoTA algorithms on LivDet-Iris 2017 databases in terms of APCER (\%), BPCER (\%) and HTER (\%) which determined by a threshold of 0.5. The \textit{Winner} in first column refers to the winner of each competition database. Bold numbers indicate the first two lowest HTERs.} 
\label{tab:livdet17_results}
\vspace{-4mm}
\end{table*}

Tab.~\ref{tab:livdet17_results} summarizes the results in terms of APCER, BPCER, and HTER on the LivDet-Iris 2017 databases. 
We evaluate the algorithms on databases provided by LivDet-Iris 2017. The evaluation and comparison on LivDet-Iris 2020 are not included due to 1) no officially offered training data, 2) not publicly available test data. Moreover, LivDet-Iris 2017 databases are designed for cross-PA and cross-database scenarios, which is still considered a challenging task. In this work, we aim to focus on the impact of the algorithm itself on PAD performance rather than the diversity of data. Consequently, to make a fair comparison with SoTA algorithms on equivalent data, we compare to the Scratch version of the D-NetPAD results \cite{DBLP:conf/icb/SharmaR20}, since Pre-trained and Fine-tuned D-NetPAD used additional data (including part of Notre Dame test data) for training. This was not an issue with the other compared SoTA methods.

\begin{table}[htbp!]
\begin{center}
\resizebox{0.4\textwidth}{!}{%
\begin{tabular}{|c|c|c|c|c|c|}
\hline
\multicolumn{2}{|c|}{\multirow{2}{*}{Database}} & \multicolumn{4}{c|}{TDR (\%) @ 0.2\% FDR} \\ \cline{3-6} 
\multicolumn{2}{|c|}{} & D-NetPAD & DenseNet & PBS & A-PBS \\ \hline
\multicolumn{2}{|c|}{Clarkson} & 92.05 & 92.89 & \textbf{94.02} & 92.35 \\ \hline
\multirow{2}{*}{Notre Dame} & K & \textbf{100} & 99.68 & 99.78 & 99.78 \\ 
 & U & 66.55 & 58.33 & 76.89 & \textbf{90.00} \\ \hline
\multicolumn{2}{|c|}{IIITD-WVU} & 29.30 & 58.97 & 69.32 & \textbf{72.00} \\ \hline
\end{tabular}}
\end{center}
\caption{Iris PAD performance reported in terms of TDR (\%) at 0.2\% FDR on the LivDet-Iris 2017 databases. K indicates known test subset and U is unknown subset. The highest TDR is in bold.}
\label{tab:livdet17_tdr}
\vspace{-4mm}
\end{table}

\begin{figure}[htbp!]
\centering{\includegraphics[width=1.0\linewidth]{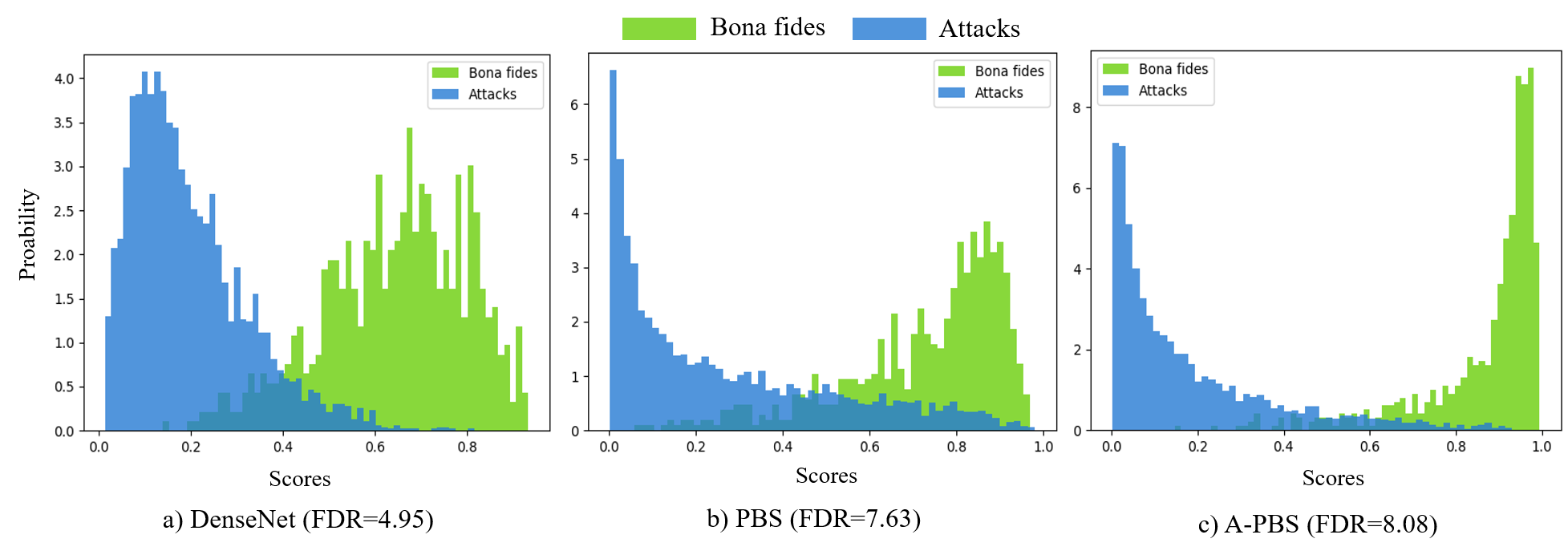}}
\caption{\textcolor{black}{Score distribution} of bona fide (green) and PAs (blue) on the IIITD-WVU test set. The histograms from left to right are produced by DenseNet, PBS, and A-PBS, respectively. The larger separability (measured by Fisher Discriminant Ratio (FDR)) and maller overlap indicate higher classification performance.}
\label{fig:score_distribution}
\vspace{-4mm}
\end{figure}

It can be observed in Tab.~\ref{tab:livdet17_results} that A-PBS architecture achieves significantly improved performance in comparison to DenseNet and also slightly lower HTER values than the PBS model in all cases. For instance, the HTER value on Notre Dame is decreased from 8.14\% by DenseNet and 4.97\% by PBS to 3.94\% by A-PBS. Although the slightly worse results on Notre Dame might be caused by the insufficient data in the training set, our PBS and A-PBS methods show significant superiority on the most challenging IIITD-WVU database. Moreover, we plot the PA score distribution of the bona fide and PAs in Fig.\ref{fig:score_distribution} for further analysis. The score distribution generated by A-PBS shows an evident better separation between bona fide (green) and PAs (blue). In addition to reporting the results determined by a threshold of 0.5, we also measure the performance of DenseNet, PBS, and A-PBS in terms of its TDR at 0.2\% FDR (to follow SoTA trends \cite{}) in Tab.~\ref{tab:livdet17_tdr}. It is worth noting that our A-PBS method obtains the highest TDR value (90.00\%) on unknown-test set in Notre Dame, while the second-highest TDR is 76.89\% achieved by PBS.  We further evaluate the generalizability of our models under cross-database scenario, e.g., the model trained on Notre Dame is tested on Clarkson and IIITD-WVU. As shown in Tab.~\ref{tab:cross_db}, the A-PBS model outperforms DenseNet and PBS in most cases, which verifying that additional spatial attention modules can reduce the overfitting of the PBS model and capture fine-grained features. Furthermore, the DenseNet and A-PBS models trained on Notre Dame even exceed the prior SoTAs when testing on the IIIT-WVU database (8.81\% HTER by DenseNet and 9.49\% by A-PBS, while the best prior SoTA achieved 11.13\% (see Tab.~\ref{tab:livdet17_results})). \textcolor{black}{
It should be noted that the Notre Dame training dataset contains only textured lens attacks while Clarkson and IIIT-WVU testing datasets comprise of both textured lens and printouts attacks, which makes this evaluation scenario partially consider unknown PAs.
In such an unknown-PAs situation, our A-PBS method achieved significantly improved results.
In general, the cross-database scenario is still a challenging problem since many D-EER values are above 20\% (Tab.~\ref{tab:cross_db}).}
\vspace{-6mm}

\begin{table*}[htbp!]
\begin{center}
\resizebox{0.87\textwidth}{!}{%
\begin{tabular}{|c|c|c|c|c||c|c|c|c||c|c|c|c|}
\hline
Trained Dataset & \multicolumn{4}{c|}{Notre Dame} & \multicolumn{4}{c|}{Clarkson} & \multicolumn{4}{c|}{IIITD-WVU} \\ \hline
Tested Dataset & \multicolumn{2}{c|}{Clarkson} & \multicolumn{2}{c|}{IIITD-WVU} & \multicolumn{2}{c|}{Notre Dame} & \multicolumn{2}{c|}{IIITD-WVU} & \multicolumn{2}{c|}{Notre Dame} & \multicolumn{2}{c|}{Clarkson} \\ \hline \hline
Metric & EER & HTER & EER & HTER & EER & HTER & EER & HTER & EER & HTER & EER & HTER \\ \hline 
DenseNet & 30.43 & 32.01 & 7.84 & \textbf{8.81} & 22.33 & 31.11 & 26.78 & 42.40 & 18.33 & 19.78 & 22.64 & 46.21 \\ \hline
PBS & 44.42 & 45.31 & 18.37 & 17.49 & 28.61 & 32.42 & 25.78 & 42.48 & \textbf{12.39} & \textbf{16.86} & 37.24 & 47.17 \\ \hline
A-PBS & \textbf{20.55} & \textbf{22.46} & \textbf{7.11} & 9.49 & \textbf{21.33} & \textbf{23.08} & \textbf{24.47} & \textbf{34.17} & 15.06 & 27.61 & \textbf{21.63} & \textbf{21.99} \\ \hline
\end{tabular}}
\end{center}
\caption{Iris PAD performance measured under cross-database scenarios and reported in terms of EER (\%) and HTER (\%). HTER is determined by a threshold of 0.5. The lowest error rate is in bold.}
\label{tab:cross_db}
\vspace{-3mm}
\end{table*}

\vspace{-2mm}
\subsection{Results on NDCLD-2013/2015 Database}

Tab.~\ref{tab:cld_results} compares the iris PAD performance of our models with five SoTA methods on NDCLD2015 and two different subsets in the NDCLD-2013 database. It can be seen from Tab.~\ref{tab:cld_results} that our A-PBS model performs the best on all databases, revealing the excellent effectiveness of a combination of PBS and attention module on textured contact lens attacks. In addition to comparison with SoTAs, we also report the TDR (\%) at 0.2\% FDR in Tab.~\ref{tab:tdr_results_nd_iiit}. Despite all three models produce similarly good results, A-PBS is still slightly better than DenseNet and PBS. The near-perfect results on NDCLD-2013/-2015 databases hint at the obsolescence and limitations of the current iris PAD databases and call for the need for more diversity in iris PAD data.

\begin{table*}[htbp!]
\begin{center}
\resizebox{0.87\textwidth}{!}{%
\begin{tabular}{|c|c|c|c|c|c|c||c|c|c|}
\hline
\multirow{2}{*}{Database} & \multirow{2}{*}{Metric} & \multicolumn{8}{c|}{Presentation Attack Detection Algorithm (\%)} \\ \cline{3-10} 
 & & LBP\cite{lbp14}  & WLBP \cite{wlbp10}  & DESIST \cite{desist16} & MHVF \cite{fusionvgg18} & MSA \cite{DBLP:conf/icb/FangDKK20, DBLP:journals/ivc/FangDBKK21} & DenseNet & PBS & A-PBS \\ \hline
\multirow{3}{*}{NDCLD15 \cite{ndcld15}} & ACPER & 6.15 & 50.58 & 29.81 & 1.92 & 0.18 & 1.58 & 1.09 & 0.08 \\ 
& BPCER & 38.70 & 4.41 & 9.22 & 0.39 & 0.00 & 0.14 & 0.00 & 0.06\\
& HTER  & 22.43 & 27.50 & 19.52 & 1.16 & 0.09 & 0.86 & 0.54 & \textbf{0.07} \\ \hline \hline
\multirow{3}{*}{NDCLD13 (LG4000) \cite{ndcld2013}} & APCER & 0.00 & 2.00 & 0.50 & 0.00 & 0.00 & 0.20 & 0.00 & 0.00  \\ 
& BPCER & 0.38 & 1.00 & 0.50 & 0.00 & 0.00 & 0.28 & 0.03 & 0.00  \\
& HTER  & 0.19 & 1.50 & 0.50 & \textbf{0.00} & \textbf{0.00} & 0.24 & 0.02 & \textbf{0.00}\\ \hline \hline
\multirow{3}{*}{NDCLD13 (AD100) \cite{ndcld2013}} & APCER & 0.00 & 9.00 & 2.00 & 1.00 & 1.00 & 0.00 & 0.00 & 0.00 \\ 
& BPCER & 11.50 & 14.00 & 1.50 & 0.00 & 0.00 & 0.00 & 0.00 & 0.00 \\
& HTER  & 5.75 & 11.50 & 1.75 & 0.50 & 0.50 & \textbf{0.00} & \textbf{0.00} & \textbf{0.00}\\ \hline
\end{tabular}}
\end{center}
\caption{Iris PAD performance of our proposed methods and existing SoTAs on NDCLD-2013/-2015 databases with a threshold of 0.5.}
\label{tab:cld_results}
\vspace{-4mm}
\end{table*}

\begin{table}[htbp!]
\def\arraystretch{1.0}
\begin{center}
\resizebox{0.33\textwidth}{!}{%
\begin{tabular}{|c|c|c|c|}
\hline
\multirow{2}{*}{Database} & \multicolumn{3}{c|}{TDR (\%) @ 0.2\% FDR} \\ \cline{2-4} 
 & DenseNet & PBS & A-PBS \\ \hline
NDCLD15 & 99.45 & 99.84 & \textbf{99.96} \\ \hline
NDCLD13 (LG4000) & 99.75 & \textbf{100} & \textbf{100} \\ \hline
NDCLD13 (AD100) & 100 & 100 & 100 \\ \hline
IIITD-CLI (Cognet) & 99.02 & \textbf{99.59} & 99.57 \\ \hline
IIITD-CLI (Vista) & 100 & 100 & 100 \\ \hline
\end{tabular}}
\end{center}
\caption{Iris PAD performance reported in terms of TDR (\%) at 0.2\% FDR on NDCLD-2013/-2015databases.}
\label{tab:tdr_results_nd_iiit}
\vspace{-4mm}
\end{table}

\vspace{-2mm}
\subsection{Results on IIITD-CLI Database}

\begin{table}[htbp!]
\begin{center}
\footnotesize
\resizebox{0.33\textwidth}{!}{%
\begin{tabular}{|c|c|c|}
\hline
PAD Algorithms & Cogent & Vista \\ \hline
Textural Features \cite{DBLP:conf/icpr/WeiQST08} & 55.53 & 87.06 \\ \hline
WLBP \cite{wlbp10} & 65.40 & 66.91 \\ \hline
LBP+SVM \cite{lbp14} & 77.46 & 76.01 \\ \hline
LBP+PHOG+SVM \cite{DBLP:conf/civr/BoschZM07} & 75.80 & 74.45 \\ \hline
mLBP \cite{iiitd_cli} & 80.87 & 93.91 \\ \hline
ResNet18 \cite{DBLP:conf/cvpr/HeZRS16} & 85.15 & 80.97 \\ \hline
VGG \cite{vgg16} & 90.40 & 94.82 \\ \hline
MVANet \cite{Gupta20} & 94.90 & 95.11 \\ \hline \hline
DenseNet & 99.37 & \textbf{100} \\ \hline
PBS & 99.62 & \textbf{100} \\ \hline
A-PBS & \textbf{99.70} & \textbf{100} \\ \hline
\end{tabular}}
\end{center}
\caption{Iris PAD performance in terms of CCR (\%) on IIITD-CLI.}
\label{tab:iiit_cld_results}
\vspace{-5mm}
\end{table}

Most of the existing works reported the results using CCR metric on IIITD-CLI database \cite{iiitd_cli, iiitd_cli_2}, we also strictly follow its experimental protocol and the experimental results are compared in Tab.\ref{tab:iiit_cld_results}. In addition, the TDR at 0.2\% FDR is reported in Tab.\ref{tab:tdr_results_nd_iiit} . The experiments are performed on Cognet and Vista sensor subsets, respectively. As shown in Tab.~\ref{tab:cld_results}, our models outperform all hand-crafted and CNN-based methods by a large margin (99.79\% on Cognet subset and 100\% on Vista subset). The near-perfect classification performance achieved by DenseNet, PBS, and A-PBS reveals that despite the improvements of deep learning models, large-scale iris PAD databases are urgently needed to be collected for future studies.

\subsection{Visualization and Analysis}
\vspace{-4mm}
\begin{figure}[htbp!]
\centering{\includegraphics[width=0.80\linewidth]{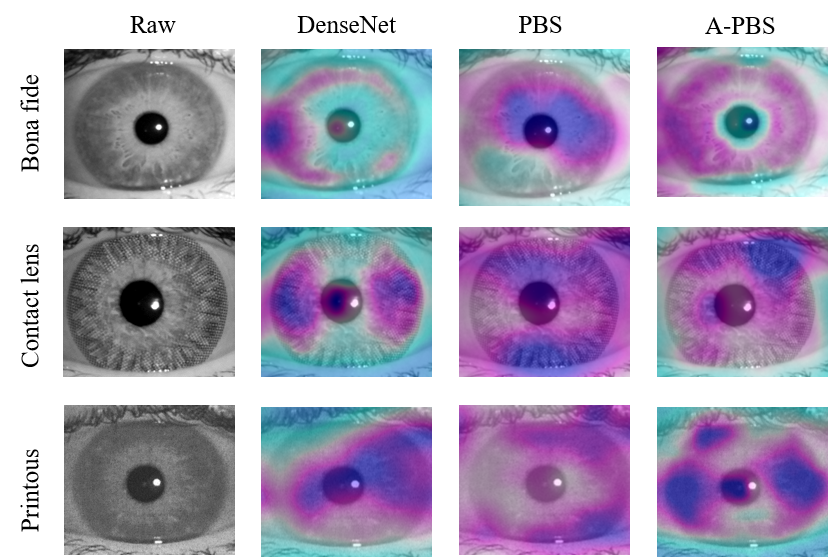}}
\caption{Score-CAM visualizations for bona fide and attack samples in the IIITD-WVU test set. \textcolor{black}{The darker the color of the region, the higher the attention on this area.} The column from left to right refers to the raw samples, maps produced by DenseNet, PBS, and A-PBS model, respectively. The first row is the bona fide samples and the left rows are textured contact lens and printouts attacks.}
\label{fig:score_cam}
\vspace{-4mm}
\end{figure}

PBS is expected to learn more discriminative features by supervising each pixel/patch in comparison with binary supervised DenseNet. Subsequently, the A-PBS model, an extended model of PBS, is hypothesized to automatically locate the important regions that carry the features most useful for making an accurate iris PAD decision. To further verify these assumptions, Score-Weighted Class Activation Mapping (Score-CAM) \cite{DBLP:conf/cvpr/WangWDYZDMH20} is used to generate the visualizations for randomly chosen bona fide and attack iris images (these images belong to the same identity) in the IIIT-WVU test set. As illustrated in Fig.~\ref{fig:score_cam}, it is clear that PBS and A-PBS models pay more attention to the iris region, while the DenseNet model seems to lose some information.  By observing the heatmaps produced by PBS, it is noticed that the attention appears to cover almost the whole iris and pupil area.  This is consistent with our assumption and expectation for PBS. Furthermore, it is clear in Fig.~\ref{fig:score_cam} that the use of the spatial attention module has enabled the model to shift more focus to the circular iris. To be specific, the attention in the corners and boundaries of the image, even including the pupil is slightly decreased.

\vspace{-3mm}

\section{Conclusion} 
\label{sec:con}

In this work, we propose a novel attention-based deep pixel-wise binary supervision (A-PBS) method for iris PAD. The proposed method aimed at 1) capture the fine-grained pixel/patch-level cues with the help of PBS, 2) find regions that the most contribute to an accurate PAD decision automatically by the attention mechanism. Extensive experiments are performed on LivDet-Iris 2017 and other three publicly available databases to verify the effectiveness and robustness of proposed A-PBS methods. The A-PBS model outperforms SoTA methods in most experimental cases including scenarios with unknown attacks, sensors, databases.

\vspace{-3mm}

{\small
\bibliographystyle{ieee}
\bibliography{main}
}

\end{document}